\documentclass[letterpaper, 10 pt, conference]{ieeeconf}  %

\IEEEoverridecommandlockouts                              %
\makeatletter
\let\NAT@parse\undefined
\makeatother

\usepackage{url}

\usepackage{makeidx}         %
\usepackage{graphicx}        %
\usepackage{multicol}        %
\usepackage{amsmath}
\usepackage{amssymb}
\usepackage{booktabs}
\usepackage{xspace}
\usepackage{color}
\usepackage[hang,flushmargin]{footmisc}
\usepackage{microtype}
\usepackage{float}
\usepackage{array}
\usepackage{multirow}
\usepackage{flushend}
\usepackage{svg}
\usepackage{makecell}
\usepackage{bm}
\usepackage{subcaption} %
\usepackage{stfloats} %

\usepackage{ifthen}
\newboolean{fourPageVersion}
\setboolean{fourPageVersion}{false}

\newboolean{anonymousSubmisson}
\setboolean{anonymousSubmisson}{true}
\setboolean{anonymousSubmisson}{false} %

\ifbool{fourPageVersion}{
    \renewcommand{\baselinestretch}{0.99}
}{}

\usepackage{cite}
\usepackage{amsmath,amssymb,amsfonts}
\usepackage{algorithmic}
\usepackage{graphicx}
\usepackage{textcomp}
\usepackage{xcolor}
\usepackage{tensor}

\usepackage{booktabs}
\usepackage{outlines}
\usepackage{caption}
\usepackage{subcaption}
\usepackage[hang,flushmargin]{footmisc}
\usepackage{stmaryrd}

\usepackage{microtype}
\usepackage{hhline}
\usepackage{multirow}
\usepackage{siunitx}
\usepackage{makecell}
\usepackage{gensymb}
\usepackage{placeins}

\usepackage{svg-extract}

\definecolor{cvprblue}{rgb}{0.21,0.49,0.74}
\usepackage[breaklinks,colorlinks]{hyperref}
\usepackage{cite}

\usepackage{xspace}

\newcommand{\etal}{\emph{et al.}\xspace}
   
\renewcommand{\[}{\begin{equation}}
\renewcommand{\]}{\end{equation}}

\newcommand{\ourName}{Real2Gen}
\newcommand{\ourNameFull}{
    Scaling Single Human Demonstrations for Imitation Learning using Generative Foundational Models
}

\newcommand{\projectsite}{
    \ifbool{anonymousSubmisson}{anonymized for review}{real2gen.cs.uni-freiburg.de}%
}

\newcommand{\absoluteDeltaIncrease}{26.6\%}

\newcommand{\filterTableOurs}{Point-E~\cite{nichol2022point} (ours)}

\newcommand{\supersubscript}[3]{\tensor[^{#1}]{#2}{_{#3}}}
\newcommand{\tf}[2]{
    \supersubscript{{#1}}{\mathbf{T}}{{#2}}
}

\newcommand{\drawFromSet}{\in}

\newcommand{\human}{h}
\newcommand{\robot}{r}

\newcommand{\primary}{p}
\newcommand{\secondary}{s}

\newcommand{\imageObservation}{o}
\newcommand{\humanObservation}{\imageObservation_\human}
\newcommand{\robotObservation}{\imageObservation_\robot}

\newcommand{\allTimesteps}{T}

\newcommand{\rgbImage}{I}
\newcommand{\depthImage}{D}

\newcommand{\primaryRGB}{\rgbImage_\primary}
\newcommand{\secondaryRGB}{\rgbImage_\secondary}

\newcommand{\mask}{m}
\newcommand{\primaryMask}{\mask_\primary}
\newcommand{\secondaryMask}{\mask_\secondary}

\newcommand{\trajectory}{J}
\newcommand{\primaryTrajectory}{\trajectory_\primary}

\newcommand{\mesh}{m}
\newcommand{\allMeshes}{M}
\newcommand{\allPrimaryMeshes}{\allMeshes_\primary}
\newcommand{\allSecondaryMeshes}{\allMeshes_\secondary}

\newcommand{\canonical}{c}
\newcommand{\metric}{m}

\newcommand{\rawMesh}{\mesh^\canonical}
\newcommand{\metricMesh}{\mesh^\metric}

\newcommand{\metricPrimaryMesh}{\metricMesh_\primary}
\newcommand{\metricSecondaryMesh}{\metricMesh_\secondary}

\newcommand{\reference}{f}
\newcommand{\referenceRGB}{\rgbImage_\reference}
\newcommand{\referenceDepth}{\depthImage_\reference}

\newcommand{\renderIndex}{r}
\newcommand{\allViews}{V}
\newcommand{\view}{v}
\newcommand{\renderAmounts}{N_\view}

\newcommand{\amountMeshes}{N_\mesh}

\newcommand{\demonstration}{d}
\newcommand{\demonstrationDataset}{D}

\newcommand{\SEThree}{SE(3)}

\newcommand{\horizonLength}{H}

\usepackage[capitalize]{cleveref}
\crefname{section}{Sec.}{Secs.}
\Crefname{section}{Section}{Sections}
\Crefname{table}{Table}{Tables}
\crefname{table}{Tab.}{Tabs.}
\crefname{algorithm}{Algo.}{Algos.}
\Crefname{algorithm}{Algorithm}{Algorithms}

\crefname{section}{Sec.}{Secs.}
\Crefname{section}{Section}{Sections}

\crefname{figure}{Fig.}{Figs.}
\Crefname{figure}{Figure}{Figures}

\crefname{table}{Tab.}{Tabs.}
\Crefname{table}{Table}{Tables}

\crefname{algorithm}{Algo.}{Algos.}
\Crefname{algorithm}{Algorithm}{Algorithms}

\crefname{subappendix}{App.}{Apps.}
\Crefname{subappendix}{Appendix}{Appendices}

\definecolor{red}{RGB}{255, 0, 0}   %
\definecolor{orange}{RGB}{255, 77, 0}   %
\definecolor{green}{RGB}{0, 128, 0}   %
\definecolor{purple}{RGB}{160, 32, 240}   %
\definecolor{lightblue}{RGB}{52, 155, 235}   %
\definecolor{darkmagenta}{RGB}{204, 51, 139} %

\def\BibTeX{{\rm B\kern-.05em{\sc i\kern-.025em b}\kern-.08em
    T\kern-.1667em\lower.7ex\hbox{E}\kern-.125emX}}

\usepackage{pifont}%
\newcommand{\cross}{\textrm{\ding{61}}}

\hypersetup{
    colorlinks=true,
    linkcolor=blue,
    filecolor=magenta,    
    citecolor=green,
    urlcolor=blue,
    pdftitle={\ourName: \ourNameFull},
}

\title{\LARGE \bf
    \ourNameFull
}

\ifbool{anonymousSubmisson}{
    \author{Anonymous Author(s)}
}{
    \author{Nick Heppert$^{*1,2}$, Minh Quang Nguyen$^{*1}$, Abhinav Valada$^{1}$
    \thanks{$^*$These authors contributed equally and are ordered alphabetically.}
    \thanks{
        \mbox{$^1$Computer} Science, University of Freiburg, Germany, 
        \mbox{$^2$Zuse} School ELIZA
    }%
    \thanks{This work was partially funded by the Carl Zeiss Foundation with the ReScaLe project. Nick Heppert is supported by the Konrad Zuse School of Excellence in Learning and Intelligent Systems (ELIZA) through the DAAD programme Konrad Zuse Schools of Excellence in Artificial Intelligence, sponsored by the Federal Ministry of Education and Research.}
    \thanks{The authors would like to thank Eugenio Chisari and Max Argus for discussion as well as Martin Büchner for measuring objects.}
    }
}

\begin{document}

\maketitle

\begin{abstract}
   Imitation learning is a popular paradigm to teach robots new tasks, but collecting robot demonstrations through teleoperation or kinesthetic teaching is tedious and time-consuming. In contrast, directly demonstrating a task using our human embodiment is much easier and data is available in abundance, yet transfer to the robot can be non-trivial. In this work, we propose \ourName{} to train a manipulation policy from a single human demonstration. \ourName{} extracts required information from the demonstration and transfers it to a simulation environment, where a programmable expert agent can demonstrate the task arbitrarily many times, generating an unlimited amount of data to train a flow matching policy. 
We evaluate \ourName{} on human demonstrations from three different real-world tasks and compare it to a recent baseline. \ourName{} shows an average increase in the success rate of $\absoluteDeltaIncrease$ and better generalization of the trained policy due to the abundance and diversity of training data. We further deploy our purely simulation-trained policy zero-shot in the real world. We make the data, code, and trained models publicly available at 
\url{https://real2gen.cs.uni-freiburg.de}.

\end{abstract}

\section{Introduction}
\label{sec:intro}

In the future, robots should be able to easily acquire new manipulation skills with minimal to no overhead in teaching them. To achieve this, imitation learning has crystallized as one of the primary paradigms for teaching a robot skills~\cite{celemin2022interactive,livanec2025designing}. Classically, imitation learning uses a dataset consisting of demonstrations of aligned robot observations as well as actions collected through \mbox{teleoperation~\cite{vonHartz2024art}}
or kinesthetic teaching~\cite{honerkamp2024zero}.
While the user operates the robot, the robot actively records its observations and actions to form a training demonstration dataset. We call these types of demonstrations robot demonstrations. One advantage of robot demonstrations is the embodiment alignment of the training data with the robot. In contrast, while collecting the data itself is not only time-consuming, \mbox{tele-operating} also requires a skilled operator. Similarly, for kinesthetic teaching, the operator needs to be physically present in the scene and, for example, can block cameras.

A recent trend aims to mitigate the tedious collection process by directly learning from demonstrations of humans in their own embodiment~\cite{welschehold2016learning, shao2021concept2robot}. We coin these types of demonstrations as human demonstrations. While human demonstrations compared to robot demonstrations are easy to collect or are even widely available on the internet~\cite{shan2020understanding}, they impose a significant challenge as the embodiment between human and robot differs. This challenge was tackled by researchers, for example, through the use of handcrafted heuristics to match a human hand to the robot gripper~\cite{papagiannis2024retrieval, lepert2025phantomt, ren2025motion} or by constraining the possible human movements in a demonstration to the robot's capabilities~\cite{xiong2021learning}. In contrast, works such as DITTO~\cite{heppert2024ditto}, ORION~\cite{zhu2024orion}, or OKAMI~\cite{okami2024} propose closing the embodiment gap by finding explicit object correspondences between the human demonstration and the robot environment at inference time, allowing for the transfer of motion in an object-centric manner.%

\begin{figure}[t]
    \centering
        \ifbool{fourPageVersion}{
            \includegraphics[width=0.8\columnwidth]{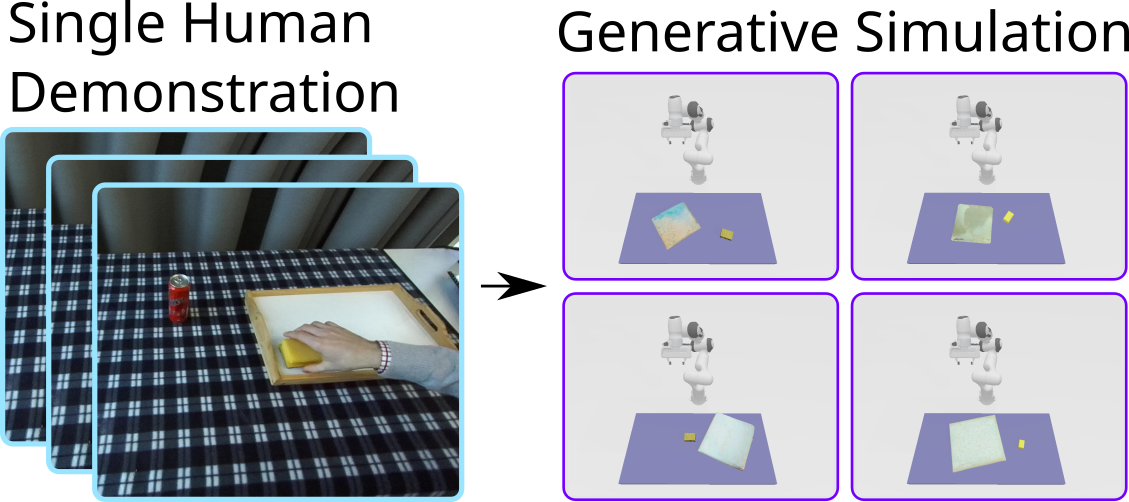}
        }{
            \includegraphics[width=0.9\columnwidth]{assets/figures/teaser}
        }
    \caption{Overview of \ourName. \ourName{} takes a single human demonstration as input and produces simulatable meshes using 3D generative foundational models, which can be used in a generative simulation setup.}
    \label{fig:teaser}
\end{figure}

In this work, we aim to combine the best of human and robot demonstrations and propose a ``Reality to Simulation'' (Real2Sim)~\cite{li2024robogsim} method called \ourName{} that processes a single human demonstration to set up a simulator that automatically allows us to collect robot demonstrations. Thus, we leverage the ease of collecting demonstrations using our human embodiment, but can train a policy directly on data in the robot embodiment. While previous Real2Sim approaches needed to scan the environment in an offline step~\cite{qureshi2024splatsim, torne2024reconciling, yu2025real2render2real}, \ourName{} uses 3D generative foundational models such as \mbox{Point-E}~\cite{nichol2022point} or \mbox{Zero-1-to-3}~\cite{liu2023zero} to directly generate 3D assets of task-relevant objects given their appearance in the human demonstration. We then use a foundational feature matcher~\cite{goodwin2022zero} to align the generated 3D assets with the human demonstration, thereby retrieving realistic scale and canonical orientation. 
Similar to other simulation works~\cite{james2019rlbench, ha2023scaling}, we first randomly sample poses of the assets and then use a scripted expert agent to generate robot demonstrations. Finally, we train a flow matching policy~\cite{chisari2024learning} on the collected robot demonstrations.\looseness=-1

We evaluate our policy in a simulation environment consisting of unseen object instances retrieved from a curated large-scale 3D object set such as Objaverse~\cite{deitke2023objaverse}. Compared to the DITTO~\cite{heppert2024ditto} baseline, we show an average success rate improvement of $\absoluteDeltaIncrease$. Lastly, we also demonstrate zero-shot transfer of our purely simulation-trained policy to a real-world setup, successfully performing the task as demonstrated by the human. While in this paper we show the capabilities of \ourName{} in an imitation learning setup, our approach also transfers without any modifications to a reinforcement learning setup.\looseness=-1

Our main contributions are as follows:
\begin{outline}
    \1 A framework to processes human demonstrations for generating robot demonstrations. %
    \1 Evaluations of mesh retrieval and generation rates as well as of scale and pose estimation methods for meshes.
    \1 Successful transfer from reality to generative simulation to reality from a single human demonstration.
    \1 Publicly available code, trained models, and data at 
    \url{https://real2gen.cs.uni-freiburg.de}.
\end{outline}
\ifbool{fourPageVersion}{
    \section{Related Work}
\label{sec:related_work}
This section reviews previous works that use human demonstrations to learn a robotic manipulation policy as well as highlights progress in the field of scaling robot learning data through generative or procedural simulations. 

\subsection{Learning from Human Demonstrations}
\label{sec:related_work:robot_learning_human_demo}
While learning from human demonstrations, rather than robot demonstrations, a prominent challenge is the gap between human and robot embodiment. Here, at the core, it is first necessary to disentangle task-relevant information from the human embodiment and second, to provide an interface to that information such that a robot can use it to execute the task using its robot embodiment. To store and access this information, a wide range of representations have been proposed, ranging from concrete object trajectories~\cite{heppert2024ditto, zhu2024orion} to affordance~\cite{papagiannis2024retrieval, srirama2024hrp} or pixel-level~\cite{xu2024flow, bharadhwaj2024track2act}, which we discuss in detail.

{\parskip=3pt
\noindent\textit{Trajectories}:}
An efficient way to represent tasks is through object-centric trajectories \cite{rana2024affordance, vonHartz2024art}. Works such as DITTO~\cite{heppert2024ditto}, ORION~\cite{zhu2024orion}, and OKAMI~\cite{okami2024} propose detecting and tracking objects in human demonstrations using foundational models. DITTO~\cite{heppert2024ditto} uses an off-the-shelf hand-object detector 
to extract relevant object masks of the task, followed by tracking correspondences throughout the scene using, e.g., LoFTR~\cite{sun2021loftr}. Using RGB-D images, the correspondences are lifted to 3D, and relative object movement is extracted between timesteps. At test-time, the trajectory is transferred using correspondences and executed using off-the-shelf grasping and motion planning libraries. ORION~\cite{zhu2024orion} follows a similar approach to extract object trajectories and for test-time inference, but instead of hand-object detectors requires a language instruction to extract relevant objects.
Additionally, ORION~\cite{zhu2024orion} segments the task demonstration into key frames. For each key frame, ORION~\cite{zhu2024orion} stores contact information and keypoints in so-called Open-World Object Graphs. 
While also leveraging foundational keypoints through KAT~\cite{di2024keypoint}, R+X~\cite{papagiannis2024retrieval} neither requires a hand-object detector nor language annotations, but instead relies on multiple demonstrations provided in a full-length video to identify common task-relevant keypoints.
{\parskip=3pt
\noindent\textit{Affordance}:}
The concept of affordance is a well-studied topic in robotics~\cite{ijcai2021p590}. Recently, researchers have begun to investigate affordance as a common representation between humans and robots, enabling easy transfer~\cite{kuang2024ram, srirama2024hrp}. Different to trajectories, affordance usually consists of a region and a simple short-horizon trajectory~\cite{bahl2023affordances}.
Similar to R+X~\cite{papagiannis2024retrieval}, the affordance-based method RAM~\cite{kuang2024ram} retrieves a demonstration from an unstructured source. 
Another body of work used human affordance as a means of pre-training an encoder for visual policy learning \cite{srirama2024hrp, bahl2023affordances, mendonca2023structured}.
{\parskip=3pt
\noindent\textit{Pixel-Level}:}
Moving away from structured and semantic interpretable representations, such as trajectories or affordances, pixel-level motion inference and subsequent transfer to the robot is another promising avenue. Here, most notably, Track2Act~\cite{bharadhwaj2024track2act} and Im2Flow2Act~\cite{xu2024flow}, follow a similar paradigm of first pre-training an open-loop 2D point inference module either on internet data~\cite{bharadhwaj2024track2act} or mixed human and robot data~\cite{xu2024flow} and second training a complementary closed-loop robot policy on robot data.

Our method \ourName{} builds upon these advances and aims to generate diverse training data using extracted object trajectories and key frames. 

\subsection{Learning through Procedural and Generative Simulation}
\label{sec:related_work:scaling_with_sim}
\begin{figure*}[t!]
    \centering
    \ifbool{fourPageVersion}{
        \includegraphics[width=0.75\linewidth]{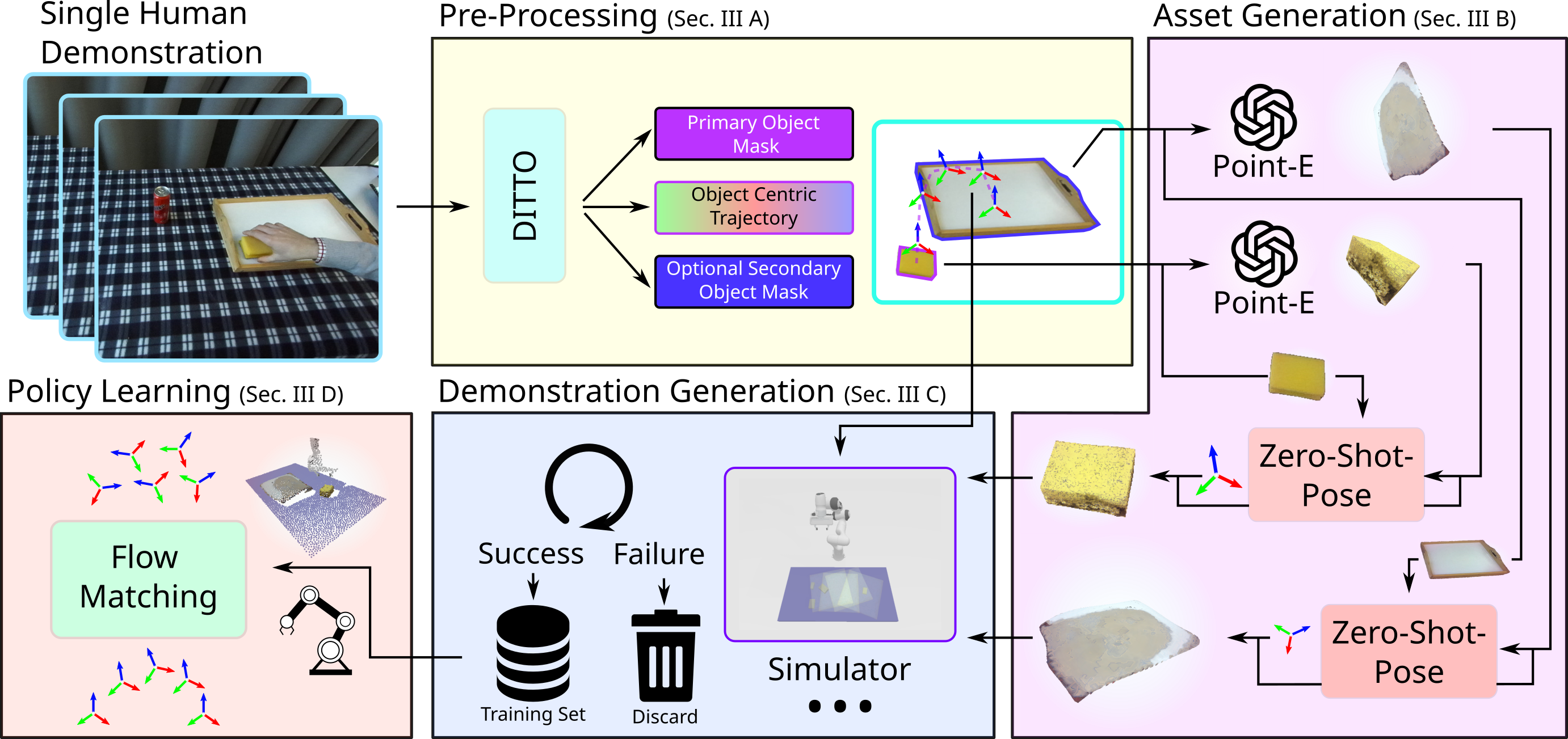}
    }{
        \includegraphics[width=0.8\linewidth]{assets/figures/overview_v3.png}
    }
    \caption{Technical approach of \ourName{}. \ourName{} uses a single human demonstration as input, consisting of a sequence of RGB-D images. We pre-process (\cref{sec:method:preprocessing}) these images using DITTO~\cite{heppert2024ditto} to retrieve a primary and, if applicable, a secondary object mask as well as an object-centric trajectory of the object. In the second step, asset generation (\cref{sec:method:datagen}), we pass object images to Point-E~\cite{nichol2022point} to generate 3D meshes in a canonical space. We then use Zero-Shot-Pose (ZSP)~\cite{goodwin2022zero} to scale and align the meshes to the human demonstration. We then use the generated meshes combined with object-centric trajectories to set up a simulation (\cref{subsec:method:demo_gen}). Using grasp and motion planning, we use the simulation to generate an expert dataset of policy rollouts. In the last step, policy learning (\cref{subsec:method:policy_learning}), we use the collected dataset to train a conditional flow matching policy~\cite{chisari2024learning}.}
    \label{fig:overview}
    \vspace{-0.3cm}
\end{figure*}
The previously discussed works either require open-loop motion planning \cite{heppert2024ditto, zhu2024orion} or robot demonstrations to fine-tune a closed-loop policy \cite{bharadhwaj2024track2act, xu2024flow}. To combine the advantages of both, the ease and determinism of motion planning and the robustness of closed-loop policy execution, automatic robot demonstration generation in real-world aligned simulations is a promising candidate \cite{james2019rlbench, ha2023scaling, yu2025real2render2real}. 

The rise of Large Language Models (LLMs) has also fueled this paradigm as LLMs can be used to, e.g., generate task-specific scenarios, success criteria, and demonstration code~\cite{ha2023scaling, wang2024gensim} given a fixed set of curated object CAD models, such as Google scanned objects~\cite{downs2022google}, to choose from.
As this limits the system in terms of object diversity, Gen2Sim~\cite{katara2024gen2sim} proposes an automated process that employs differential rendering to generate 3D assets from any 2D image of a singulated object. The images can either stem from the robot's workspace, the web, or an image generation model, e.g., Stable Diffusion \cite{rombach2021highresolution}. To ground the generated assets with realistic sizes and masses Gen2Sim~\cite{katara2024gen2sim} uses a Vision Language Model (VLM). Another LLM then generates meaningful tasks for given scenes as well as rewards to train a reinforcement learning agent. Given a live observation GRS~\cite{zook2024grs} uses a VLM to describe and match all objects in their 3D asset set to the shown scene. Unlike Gen2Sim~\cite{katara2024gen2sim}, the authors assume that the 3D asset set is already provided, e.g., through Objaverse~\cite{deitke2023objaverse}. 

RoboGen~\cite{wang2024robogen} aims to combine all these previous approaches by allowing scene generation with either assets generated from 3D foundational models or from a large dataset, such as Objaverse~\cite{deitke2023objaverse}. Similarly as Gen2Sim~\cite{katara2024gen2sim} and GRS~\cite{zook2024grs}, RoboGen~\cite{wang2024robogen} verifies meshes using a VLM. Unlike other works, they also tackle long-horizon tasks by allowing an LLM to decide whether sub-skills should be learned through reinforcement learning, gradient-based trajectory optimization, or motion planning. RialTo~\cite{torne2024reconciling} goes a slightly different path by not relying on any of these expert ways but rather uses a trained policy trained on real robot demonstrations and only saves successful executions to form a final dataset. RialTo~\cite{torne2024reconciling} generates their simulation environment through scanning the real scene and annotating it.\looseness=-1

Different from these works, \ourName{} directly uses generative 3D foundational models to generate 3D CAD models aligned with the human demonstration, as well as provides an automatic and robust alignment process of the generated CAD models to the demonstration.
As a result, \ourName{} needs no additional manual curation of meshes or scenes, nor does \ourName{} require text-input that describes the task or iterative code generation of LLMs.

}{
    
}

\section{Technical Approach}
\label{sec:technical_approach}
\ifbool{fourPageVersion}{
    \begin{figure*}[t!]
    \centering
    \ifbool{fourPageVersion}{
        \includegraphics[width=0.75\linewidth]{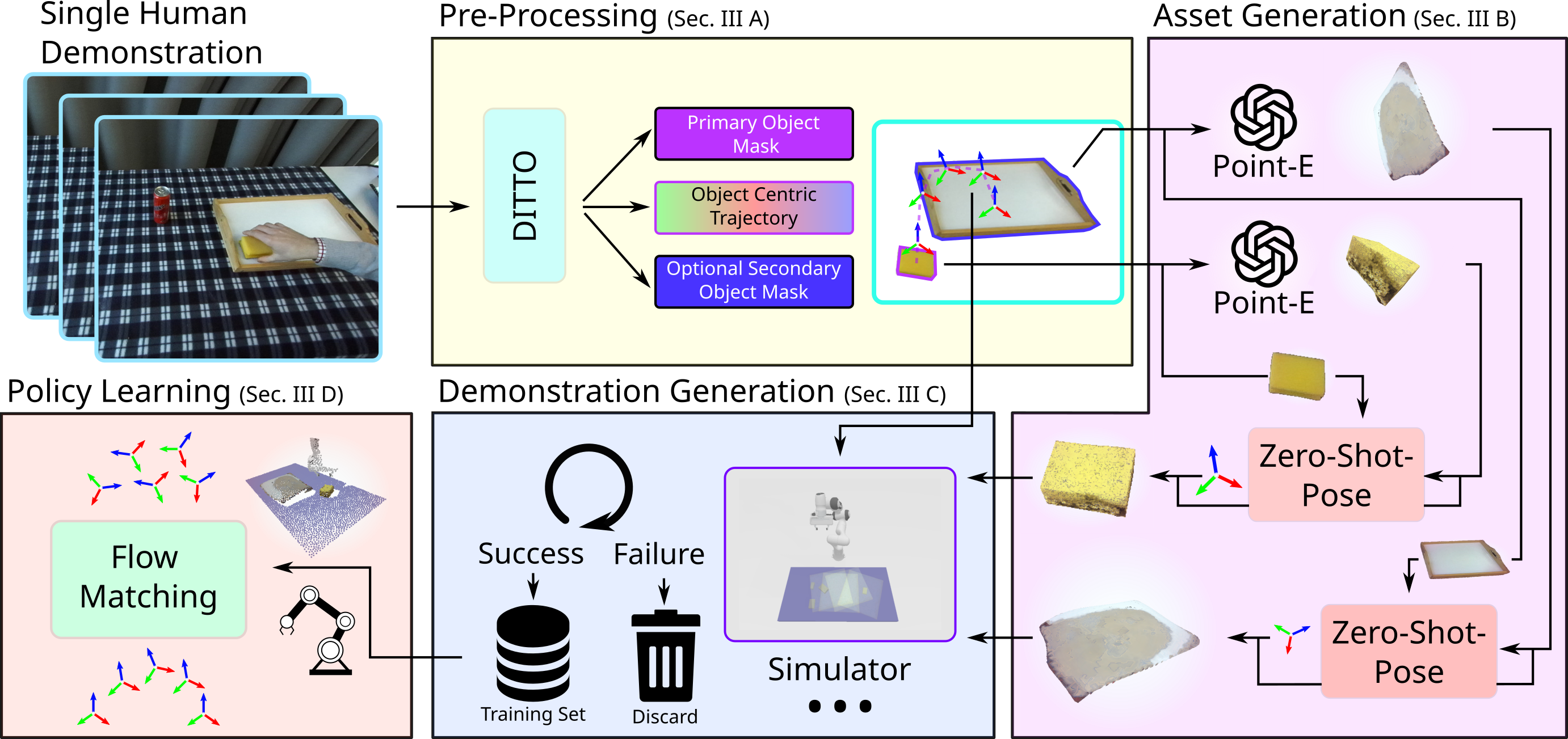}
    }{
        \includegraphics[width=0.8\linewidth]{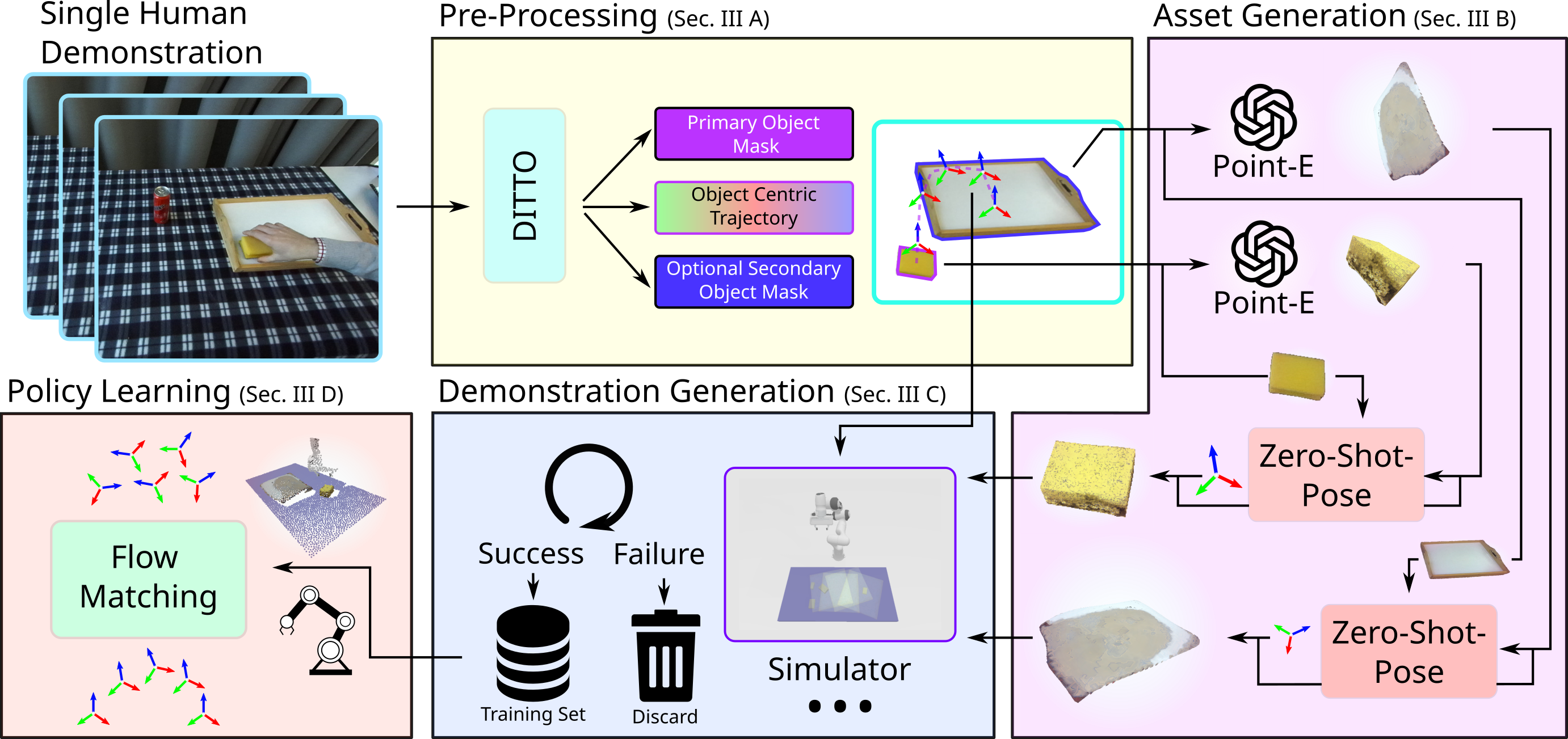}
    }
    \caption{Technical approach of \ourName{}. \ourName{} uses a single human demonstration as input, consisting of a sequence of RGB-D images. We pre-process (\cref{sec:method:preprocessing}) these images using DITTO~\cite{heppert2024ditto} to retrieve a primary and, if applicable, a secondary object mask as well as an object-centric trajectory of the object. In the second step, asset generation (\cref{sec:method:datagen}), we pass object images to Point-E~\cite{nichol2022point} to generate 3D meshes in a canonical space. We then use Zero-Shot-Pose (ZSP)~\cite{goodwin2022zero} to scale and align the meshes to the human demonstration. We then use the generated meshes combined with object-centric trajectories to set up a simulation (\cref{subsec:method:demo_gen}). Using grasp and motion planning, we use the simulation to generate an expert dataset of policy rollouts. In the last step, policy learning (\cref{subsec:method:policy_learning}), we use the collected dataset to train a conditional flow matching policy~\cite{chisari2024learning}.}
    \label{fig:overview}
    \vspace{-0.3cm}
\end{figure*}
 
}{
}
In this section, we detail our \ourName{} approach. We start by describing the pre-processing procedure in \cref{sec:method:preprocessing}, how we generate 3D assets in \cref{sec:method:datagen} and robot training data in \cref{subsec:method:demo_gen}, and lastly detail our policy learning approach in \cref{subsec:method:policy_learning}. An overview of \ourName{} approach is presented in \cref{fig:overview}.

\subsection{Pre-Processing Human Demonstrations}
\label{sec:method:preprocessing}
The input to \ourName{} is a single human demonstration consisting of a sequence of~$\allTimesteps$ RGB-D images~$\humanObservation$. We pre-process the sequence~$\humanObservation$ using DITTO~\cite{heppert2024ditto}, an object trajectory extractor, but other options such as ORION~\cite{zhu2024orion} can also be used. Similarly to these methods, we assume there is a primary object~$\primary$ moving, either in free-form or depending on another secondary goal object~$\secondary$. The goal of the pre-processing step is to extract segmentation masks of all relevant objects, i.e.,~$\primaryMask$ and~$\secondaryMask$ if applicable, in the first RGB image~$\rgbImage_0$ as well as the object-centric trajectory of the primary object~$\primaryTrajectory$ consisting of relative poses.
By segmenting the first RGB image~$\rgbImage_0$ using~$\primaryMask$ and~$\secondaryMask$ masks, we get unobstructed object-centric RGB reference images~$\primaryRGB$ and~$\secondaryRGB$ of our primary and secondary objects, respectively, before the human interacts with them.

\subsection{Asset Generation}
\label{sec:method:datagen}
Given one of the reference object RGB images~$\referenceRGB$, i.e., either~$\primaryRGB$ or~$\secondaryRGB$ if applicable, we use an off-the-shelf 3D generative foundational model, specifically Point-E~\cite{nichol2022point} combined with marching cubes, to generate a raw 3D mesh~$\rawMesh$. We decided on this combination because of its ease of use, but other models such as Zero-1-To-3~\cite{liu2023zero}, Stable-Dreamfusion~\cite{tang2022stable}, or Shape-E~\cite{jun2023shap} can also be used. As these models return object meshes in a canonical non-metric frame, they are not directly usable in a robot simulation. Although VLMs were previously used to verify the mesh and guess a metric dimension~\cite{katara2024gen2sim}, we use our provided human demonstration and match the generated mesh to it. As our generated mesh and the object in the human demonstration share semantic similarities but are not the exact instance, methods like FoundationPose~\cite{wen2024foundationpose} are not applicable here. Instead, we propose to use a 
foundational feature matcher, Zero-Shot-Pose (ZSP)~\cite{goodwin2022zero}, to first generate correspondences and then subsequently align the correspondences through a 7-DoF affine transformation with a closed form solution \cite{umeyama1991least}. We detail this approach in the following. 

Given a generated mesh~$\rawMesh$ from our 3D generative foundational model, we render~$\renderAmounts$ views~$\allViews$ of it using spherical Fibonacci sampling~\cite{gonzalez2010measurement} to sample the polar and azimuthal angles, similar to what is done by 
Giammaorino~\etal\cite{di2024learning}.
The Fibonacci sampling procedure ensures maximum diversity in views compared to densely sampling the polar and azimuthal angles. For each view~$\view_\renderIndex = \{\rgbImage_\renderIndex, \depthImage_\renderIndex\}$, we render an RGB image~$\rgbImage_\renderIndex$ and a depth image~$\depthImage_\renderIndex$. We collectively give all views~$\allViews$ and the initial reference image~$\referenceRGB$ to ZSP~\cite{goodwin2022zero}. ZSP~\cite{goodwin2022zero} then first extracts descriptors for each view's RGB image~$\rgbImage_\renderIndex$, matches these descriptors to the reference image~$\referenceRGB$ to select the view~$\hat{\view}$ with the top-K similar descriptors. Second, ZSP~\cite{goodwin2022zero} uses the most similar descriptors as correspondences and projects them in 3D using the selected views~$\hat{\view}$ depth image~$\hat{\depthImage}$, and the reference depth image~$\referenceDepth$. Lastly, ZSP~\cite{goodwin2022zero} solves a closed-form seven degree of freedoms (7-DoF) least squares problem~\cite{umeyama1991least} to estimate a full affine transformation~$\tf{\canonical}{\metric}$ to transform the canonical mesh~$\rawMesh$ to a metric mesh~$\metricMesh$ of which the up-orientation matches the human demonstration. Despite using the top-K similar descriptors across all views, there might be outliers present. Thus, the estimate is done multiple times using RANSAC. 
This process can be repeated arbitrarily many times to continuously generate new and unseen meshes usable in a robotic simulation. In practice, we generate a fixed amount of~$\amountMeshes$ meshes 
$\allMeshes$ 
before continuing with our demonstration generation. We additionally verify all matches before proceeding, which, with improvements in 3D generative foundational models, will become obsolete.
Throughout our experiments, we sampled~$\renderAmounts = 41$ views in the upper hemisphere and used the top-$30$ descriptors in ZSP~\cite{goodwin2022zero}.\looseness=-1

\subsection{Demonstration Generation}
\label{subsec:method:demo_gen}
After we generate a set of meshes~$\allPrimaryMeshes$ for the primary object and if applicable, a set~$\allSecondaryMeshes$ for the secondary object, we use them in a simulation environment to generate large-scale robot data. Specifically, we use SAPIEN~\cite{Xiang_2020_SAPIEN} as our simulator with a Franka Panda Robot mounted on a table. To observe the scene, we set up an external RGB-D camera as well as attach an RGB-D wrist camera to the robot. We assign a default density of $\qty{1000}{\kilo\gram\per\meter\cubed}$ to generated meshes.

To generate a robot demonstration, we first sample random meshes, one for the primary object~$\metricPrimaryMesh \drawFromSet \allPrimaryMeshes$ and if needed, one for the secondary object~$\metricSecondaryMesh \drawFromSet \allSecondaryMeshes$. We then sample random poses, constraining the meshes to be on the table and randomly rotated around the up-axis. Next, using our privileged simulation information, we analytically calculate grasps on the primary object mesh using antipodality constraints\footnote{In practice, this process was done in a pre-processing step for each mesh and stored. At demonstration generation time, we only perform collision checks to filter out grasps.} and pick a random grasp. At this point, we explicitly differentiate between tasks only moving the primary object and tasks with an additional secondary object. For tasks that only involve a primary object, we apply the extracted object-centric trajectory~$\primaryTrajectory$ to the grasp. For tasks consisting of an additional secondary object, we constrain the primary object to be placed in the center of the secondary object approached through a bottleneck pose above the table. 
While directly using the object-centric trajectory is also possible, in practice though, we observed higher generation failure rates, prolonging the generation unnecessary.
To finally execute the task, we use a motion planner\footnote{\url{https://motion-planning-lib.readthedocs.io/latest/}}. During execution, we record all observations~$\robotObservation$ and the robots proprioceptive states as end-effector poses. After the roll-out is completed, using the proposed pose error metric by GraspNet~\cite{fang2020graspnet}, we compare the distance between the actual pose of the primary object and the expected pose to determine whether the roll-out was a success or failure. Only successful roll-outs~$\demonstration$ are added to our demonstration dataset~$\demonstrationDataset$.

\subsection{Policy Learning}
\label{subsec:method:policy_learning}
The last component of \ourName{} consists of learning a manipulation policy given our previously generated demonstration dataset~$\demonstrationDataset$. As our policy model, we chose to use PointFlowMatch~\cite{chisari2024learning}, a flow matching-based imitation learning method. Similar to PointFlowMatch~\cite{chisari2024learning}, our policy learns to move a Gaussian distribution to a target distribution being our ground truth actions where the process is conditioned on the current robot observations. The robot observations are the last two point clouds encoded using a PointNet~\cite{qi2017pointnet} encoder as well as the last two robot proprioceptive states, in our case, the end-effector pose in~$\SEThree$.%
\ifbool{fourPageVersion}{
    We use an aggregated point cloud from the wrist camera and one external camera by concatenating the points and randomly downsample it to a fixed size.
}{
    Different from PointFlowMatch, we only use a partial point cloud from the wrist and an external camera, and do not aggregate point clouds across multiple cameras observing the scene. 
}%
Our inferred actions are future end-effector poses with a fixed horizon length~$\horizonLength$. See \cref{app:sec:policy_hps} for our hyperparameter setup.

\section{Experimental Evaluation}
\label{sec:experiments}
In our experiments, we quantitatively evaluate how well \ourName{} can transfer the human demonstration to a robot compared to a DITTO~\cite{heppert2024ditto} baseline. We also investigate the difference in quality and effort when using a 3D generative foundational model over a large-scale object dataset, such as Objaverse~\cite{deitke2023objaverse}. Further, we compare our proposed matching procedure to a VLM-based approach. Last, we show zero-shot applicability of \ourName{} on a real-robot system.

\subsection{Quantitative Full Pipeline Evaluation}
For our evaluation, we selected three tasks from DITTO~\cite{heppert2024ditto}, \texttt{Sponge on Tray}, \texttt{Coke on Tray}, and \texttt{Paperroll upright} (see \cref{app:sec:task_overview} for further details), and, following DITTO~\cite{heppert2024ditto}, a single human demonstration for each task.
We generate five meshes and 800 demonstrations. We use a simulator with the same fixed seeds across all evaluations for a fair and reproducible comparison. We evaluate \ourName{} against DITTO~\cite{heppert2024ditto} baselines 
as well as ablate \ourName{}. 

{\parskip=3pt
\noindent\textit{Evaluation Setup}:}
To set up and align our evaluation simulator with our tasks, we use unseen realistic 3D meshes from Objaverse~\cite{deitke2023objaverse}. All meshes were manually checked and scaled to have reasonable geometry and realistic sizes. 
For the \texttt{Sponge on Tray} and \texttt{Can on Tray}-task, we select five sponge, five can, and seven tray meshes. For the \texttt{Paperroll upright}-task, we select five paperroll meshes.
As for generating robot demonstrations (see \cref{sec:method:datagen}), during each evaluation episode, we randomly select meshes according to the task and spawn them at random poses. 
To perform the evaluation, we select three random seeds. For each seed, we evaluate each method 100 times and record the successes of the roll-outs. 
We evaluate the successes as described in \cref{subsec:method:demo_gen}. As done in \cite{chisari2024learning, funk2024actionflow}, we report the mean and standard deviation across all seeds.

\begin{table}
    \centering
    \resizebox{\linewidth}{!}{%
    \begin{tabular}{r|ccc|c}
        \toprule
            Method
          & \makecell{\texttt{Sponge}\\\texttt{on Tray}}
          & \makecell{\texttt{Coke on}\\\texttt{Tray}}
          & \makecell{\texttt{Paperroll}\\\texttt{upright}}
          & \makecell{Mean\\SR ($\uparrow$)} \\
        \midrule
        DITTO~\cite{heppert2024ditto} 
            & $6.3^{\pm 2.1}$ & $26.0^{\pm 3.6}$ & $0.3^{\pm 0.5}$ & $10.9^{\pm 2.1}$ \\
        DITTO~\cite{heppert2024ditto} w/ ZSP~\cite{goodwin2022zero} 
            & $4.3^{\pm 1.2}$ & $19.7^{\pm 3.8}$ & $0.7^{\pm 0.6}$ & $8.2^{\pm 1.8}$ \\
        \ourName{} (ours) 
            & $\textbf{41.3}^{\pm 4.5}$ & $\textbf{46.3}^{\pm 6.4}$ & $\textbf{25.0}^{\pm 1.0}$ & $\textbf{37.5}^{\pm 3.0}$ \\
        \bottomrule
    \end{tabular}
    }
    \caption{Performance comparison of \ourName{} against different baseline methods. We evaluate all methods on three different tasks and report per-task success rate as well as overall mean success rate $\lbrack \% \rbrack$ \mbox{($\uparrow$)}.}
    \label{tab:main_comparison}
\end{table}

{\parskip=3pt
\noindent\textit{Baseline Comparisons}:}
We compare \ourName{} against two variants of DITTO~\cite{heppert2024ditto}. First, the original variant using LoFTR~\cite{sun2021loftr} to match the live observation to the demonstration. This is an unfair comparison as DITTO~\cite{heppert2024ditto} was designed to be faced with the same object instance as in the demonstration. Thus, we extend DITTO~\cite{heppert2024ditto} by replacing the LoFTR~\cite{sun2021loftr} matching step with ZSP~\cite{goodwin2022zero} which should enable DITTO~\cite{heppert2024ditto} to work across instances from the same category. 

The results of our experiment are reported in \cref{tab:main_comparison}. 
When comparing \ourName{} to DITTO~\cite{heppert2024ditto} with and without ZSP~\cite{goodwin2022zero}, we see that the overall performance of \ourName{} is better. We assume the difference stems from DITTO~\cite{heppert2024ditto} having only a single test-time image available, which increases difficulty for matching. To our surprise, the original DITTO~\cite{heppert2024ditto} baseline outperforms the variant with ZSP~\cite{goodwin2022zero}.
Additionally, we observe that grasp and motion planning fail quite often, aligning with the reported results in DITTO~\cite{heppert2024ditto}, whereas our learned policy produces more robust results.

{\parskip=3pt
\noindent\textit{Ablation Studies}:}
We additionally perform ablation studies of \ourName{} to study the effect of the number of generated meshes and demonstrations.%
\ifbool{fourPageVersion}{
    Results are reported in the appendix in \cref{app:sec:ablation}.
}{
    \begin{figure}
    \centering
    \includegraphics[width=0.8\linewidth]{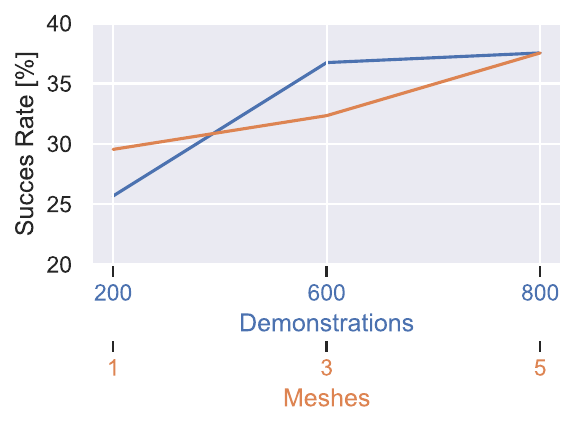}
    \caption{Results of Ablation Study. We show the average success rate $\lbrack \% \rbrack$ \mbox{($\uparrow$)} across all tasks. We either vary the number of demonstrations while using five meshes or we vary the number of meshes using 800 demonstrations.}
    \label{fig:ablation:demos_and_meshes}
\end{figure}

    Specifically, we change the number of demonstrations from 800 to 600 and 200, while still using five meshes, as well as changing the number of meshes to three and one with 800 demonstrations. We visualize the results in \cref{fig:ablation:demos_and_meshes}. As one would expect, the more meshes and demonstrations, the better. Nonetheless, the average performance increase diminishes rather soon when going from 600 to 800 ($1.1\%$) demonstrations and three to five meshes ($5.2\%$), respectively.
}

\subsection{Comparison of Mesh Generation}
\label{sec:mesh_filtering}
\ifbool{fourPageVersion}{
    \input{assets/tables/filter_effort_small}
}{
    \begin{table*}[!t]
    \centering
    \begin{tabular}{cc|ccc|c}
        \toprule
        Category & Source 
            & \makecell{Available\\Meshes} & 100 Mesh Pre-Selection & Matching Successful$^\cross$ & \ \makecell{Matching Successful\\and Task Relevant$^\cross$} \\
        \midrule\midrule
        \multirow{3}{*}{Sponge} & \filterTableOurs & $\infty$ & Random & $99\%$ & $61\%$ \\
            & \multirow{2}{*}{Objaverse~\cite{deitke2023objaverse}} & \multirow{2}{*}{351} & Most viewed & $58\%$ & $3\%$ \\
            &   &   & Random & $57\%$ & $7\%$ \\
        \midrule
        \multirow{3}{*}{Tray} & \filterTableOurs & $\infty$ & Random & $35\%$ & $18\%$ \\ 
            & \multirow{2}{*}{Objaverse~\cite{deitke2023objaverse}} & \multirow{2}{*}{278} & Most viewed & $21\%$ & $10\%$ \\
            &   &   & Random & $14\%$ & $5\%$ \\
        \midrule
        \multirow{2}{*}{Coke Can} & \filterTableOurs & $\infty$ & Random & $67\%$ & $39\%$ \\
            & {Objaverse~\cite{deitke2023objaverse}} & {41} & All & $82\%$ & $73\%$ \\
        \midrule
        \multirow{2}{*}{Paper Roll} & \filterTableOurs & $\infty$ & Random & $82\%$ & $62\%$ \\
            & {Objaverse~\cite{deitke2023objaverse}} & {20} & All & $45\%$ & $35\%$ \\
        \midrule\midrule
        \multirow{3}{*}{All} & \filterTableOurs & $\pmb{\infty}$ & Random & $\textbf{71\%}$ & $\textbf{45\%}$ \\
            & \multirow{2}{*}{Objaverse~\cite{deitke2023objaverse}} & \multirow{2}{*}{690} & Most viewed or All & $52\%$ & $30\%$ \\
            &   &   & Random or All & $50\%$ & $28\%$ \\
    \bottomrule
    \end{tabular}
    \caption{Comparison of Mesh Generation. We compare the number of resulting meshes when using our proposed generative way vs. using a large-scale object dataset.
        \textit{$^\cross$We report the total percentage from the 100 selected meshes, or if fewer than 100 meshes are available from all available ones.}
    }
    \label{tab:filter_effort}
    \vspace{-0.3cm}
\end{table*}

}

To highlight the effectiveness of our proposed automatic mesh generation, we compare the human effort needed to perform manual mesh retrieval from a large database, as performed, for instance, in Scaling Up And Distill Down~\cite{ha2023scaling}. To have representative results, we set the goal to generate 100 object meshes. For \ourName{}, this process is straightforward, and we run the asset generation a hundred times. For the object set comparison, we query Objaverse~\cite{deitke2023objaverse} to retrieve meshes with the tag of either \texttt{sponge}, \texttt{coke/coca can}, \texttt{tray}, or \texttt{paperroll}. If there are more than a hundred meshes available, we evaluate using the hundred most viewed meshes and a hundred random meshes. We additionally report the total meshes available.
We then apply the same matching procedure as for the generated meshes. 
Lastly, we manually go through all of the generated and retrieved object set meshes and classify whether they are task relevant. For the retrieved meshes, this is particularly important as they are not anchored to the human demonstration, e.g., for \texttt{sponge}, sometimes the cartoon character \texttt{Sponge Bob} is returned.
We report the results in \ifbool{fourPageVersion}{\cref{tab:filter_effort_small}}{\cref{tab:filter_effort}}. Overall, compared to previous retrieval-based methods \cite{ha2023scaling}, our proposed way to generate meshes through a generative model results in almost 50\% more available meshes on average, while also being able to generate an infinite amount.

\subsection{VLMs as Size and Pose Estimators}
\label{sec:vlm_estimators}
We further evaluate the quality and robustness of using a VLM to estimate physical properties as proposed in Gen2Sim~\cite{katara2024gen2sim}, as opposed to using our introduced foundational matching procedure (see \cref{sec:method:datagen}). We can not assume a known one-dimensional scale or canonical orientation for the generated meshes from \ourName{} nor for the Objaverse meshes. Thus, we evaluate inference performance for both physical properties in this experiment.

{\parskip=3pt
\noindent\textit{Data}:} We evaluate on a subset of 89 of all resulting matchable and semantically meaningful meshes from  \cref{sec:mesh_filtering} (essentially all last columns for each row in \cref{tab:filter_effort})\footnote{As for some sets the resulting amount of meshes is small, we test on all available meshes or on a maximum of ten meshes per set to prevent a strong bias in our results while still having sufficient and diverse meshes for statistic significance.}. Both our proposed matching and the by Gen2Sim~\cite{katara2024gen2sim} inspired VLMs receive 10 rendered images as input, sampled from the full Fibonacci sphere as described in \cref{sec:method:datagen}.

{\parskip=3pt
\noindent\textit{Baselines}:} We compare our deterministic matching approach
to a VLM baseline, as done in Gen2Sim~\cite{katara2024gen2sim}. For a fair comparison, we provide the VLM with the same inputs that the matching uses. For the rendered RGB images, this process is straightforward. Additionally, we provide the dimensions of the unscaled mesh when loaded from the file. To ground the prediction in the human demonstration, we compare providing the VLM with the full image $\rgbImage_0$ or only with a segmented and cropped image of the object $\referenceRGB$. 
\noindent\begin{minipage}[t]{\columnwidth}
    \centering
    \begingroup
    \centering
    \includegraphics[width=1.0\linewidth]{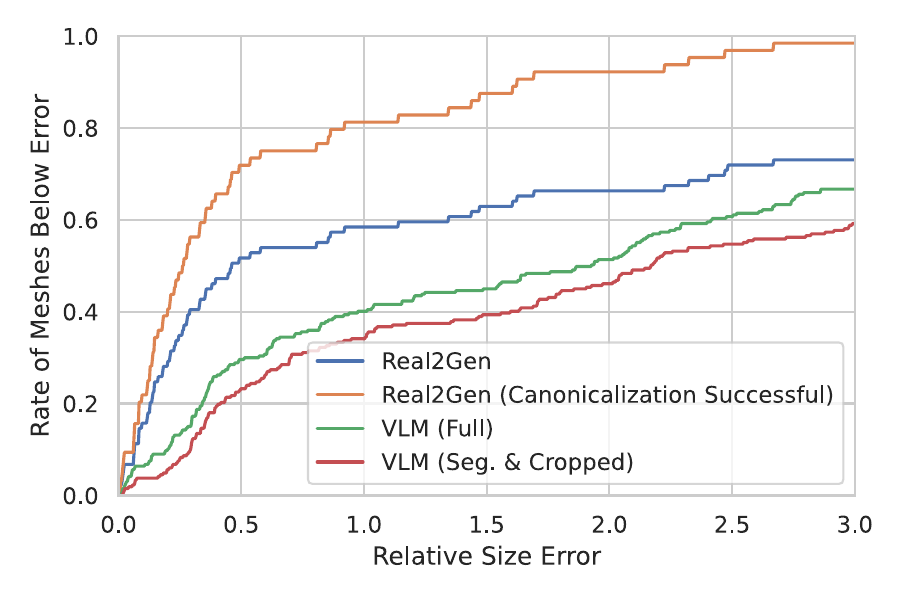}
    \vspace{-2em}
    \captionof{figure}{Precision Curve for Scaling Factor. We plot the percentage of meshes below the relative size error ranging.
    }
    \label{fig:scaling_factor}
    \endgroup
  \vspace*{1\floatsep}
    \begingroup
        \resizebox{\columnwidth}{!}{%
        \begin{tabular}{cc|cc}
            \toprule
            Method & Reference Image & \makecell{Canonicalization\\Succesful \mbox{($\uparrow$)}} & \makecell{Scaling\\mAP \mbox{($\uparrow$)}} \\
            \midrule
            \multirow{1}{*}{\ourName{} (ours)} 
                & Seg. \& cropped & \textbf{71.9\%} & 0.586 \\
            \multirow{2}{*}{VLM (Gen2Sim \cite{katara2024gen2sim})} 
                & Full & 21.3\$ & 0.380 \\
                & Seg. \& cropped & 26.9\% & 0.438 \\
            \bottomrule
            \multicolumn{4}{c}{Only Successful Canonicalization}\\ 
            \midrule
            \multirow{1}{*}{\ourName{} (ours)} 
                & Seg. \& cropped & N/A & \textbf{0.810} \\
            \multirow{2}{*}{VLM (Gen2Sim \cite{katara2024gen2sim})} 
                & Full & N/A & 0.297 \\
                & Seg. \& cropped & N/A & 0.401 \\             
            \bottomrule
        \end{tabular}}
        \captionof{table}{Comparison of Physical Property Inference. We compare our proposed matching against a VLM to infer the pose canonicalization success as well as calculate the mAP for relative errors between 0 and 3.}
        \label{tab:size_guessing}
    \endgroup
\end{minipage}

\noindent Lastly, we also provide both VLM variants with the point cloud dimensions of the segmented object in the reference depth image $\referenceDepth$. Given this information, the VLM has to infer a scaling factor for the mesh as well as determine the closest view of the rendered images to the reference image.
As VLMs are non-deterministic, to further test the robustness, we query both variants three times. 

Our matching process requires no modification for this experiment. If it fails, we will set the scaling result to identity.

{\parskip=3pt
\noindent\textit{Metrics}:} For size comparison, we measure real-world dimensions for each object category (see \cref{fig:real_object_sizes}). Given these measurements, we optimize for the closest scaling factor $\delta$ of each mesh instance to the real-world reference size. The optimized scaling factor $\delta$ for each mesh serves as our ground truth value, and we calculate the relative distance $(\hat{\delta} - \delta) / \delta$ to the inferred scaling factor $\hat{\delta}$. To compare the canonicalization of the object, we rely on proxies for both methods. For the matching approach, we directly use the information on whether the matching procedure succeeded. For the VLMs, we consider the canonicalization as successful if all three queries result in the same decision for the closest view.\looseness=-1

{\parskip=3pt
\noindent\textit{Results}:} Overall, we observe that for dimensions estimation as well as object canonicalization, the matching procedure outperforms the VLM baseline. For the scaling factor, we plot the precision curve for relative size error thresholds ranging from 0 to 3 in \cref{fig:scaling_factor}. In \cref{tab:size_guessing} we compare the success of mesh canonicalization to the reference image as well as calculate the area under the scale factor precision curve.

\subsection{Real-World Robotic Experiments}
\label{sec:exp:real}
\begin{figure}
    \centering
    \begin{subfigure}{0.9\linewidth}
        \centering
        \includegraphics[width=0.8\linewidth]{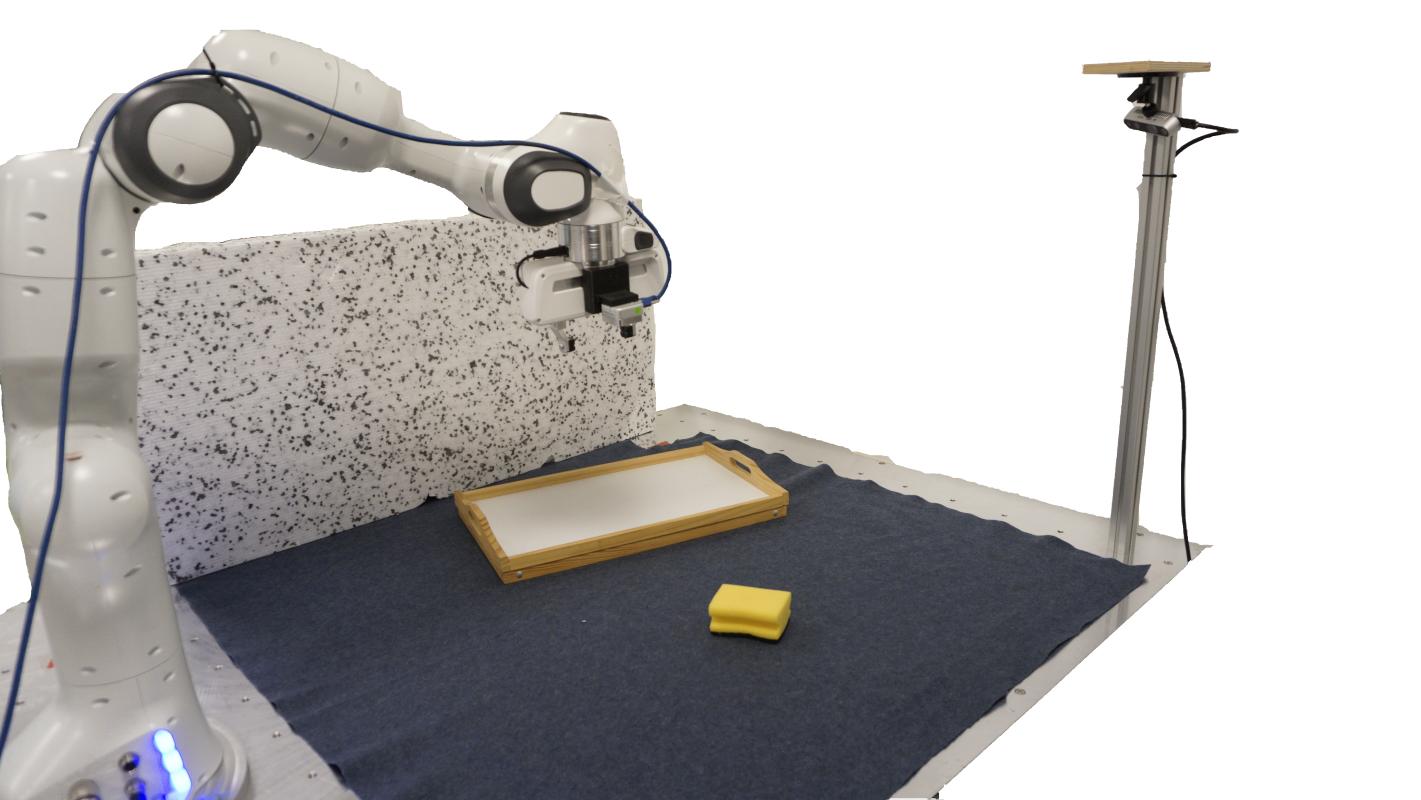}
        \caption{Real-World Setup. As in simulation, we mount a wrist camera to the robot and place an external camera overlooking the scene.}
        \label{subfig:real_world_setup}
    \end{subfigure}
    \vskip0.4\baselineskip %
    \begin{subfigure}[t]{0.49\linewidth}
        \centering
        \includegraphics[width=0.47\linewidth]{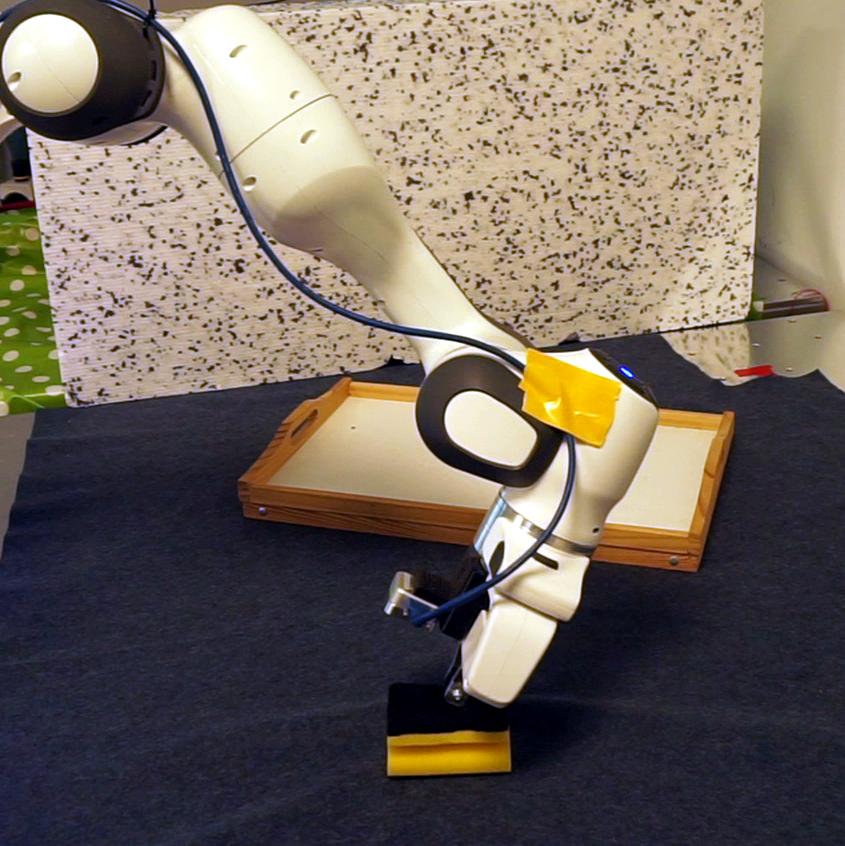}
        \hfill
        \includegraphics[width=0.47\linewidth]{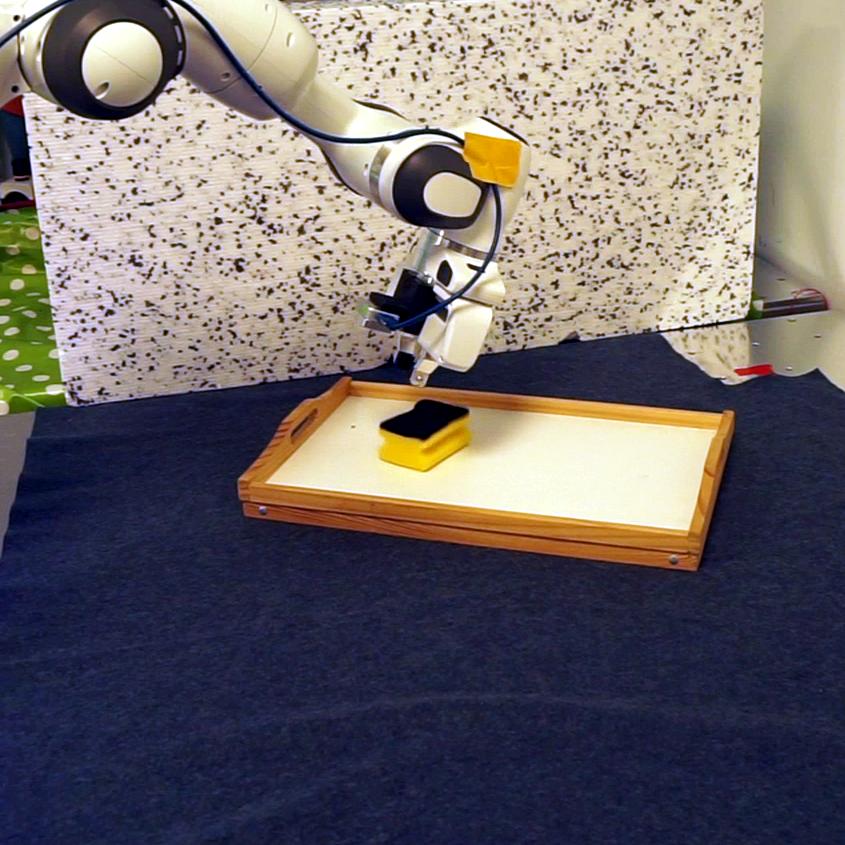}
        \caption{Successful Task Execution.}
        \label{subfig:real_world_success}
    \end{subfigure}
    \hfill
    \begin{subfigure}[t]{0.49\linewidth}
        \centering
        \includegraphics[width=0.47\linewidth]{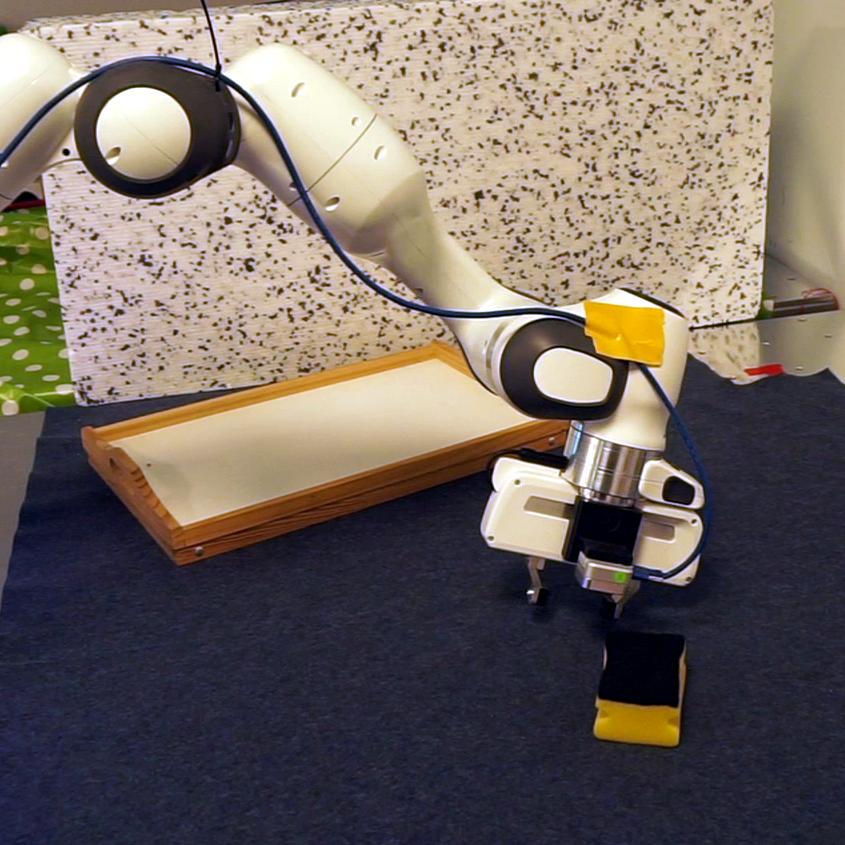}
        \hfill
        \includegraphics[width=0.47\linewidth]{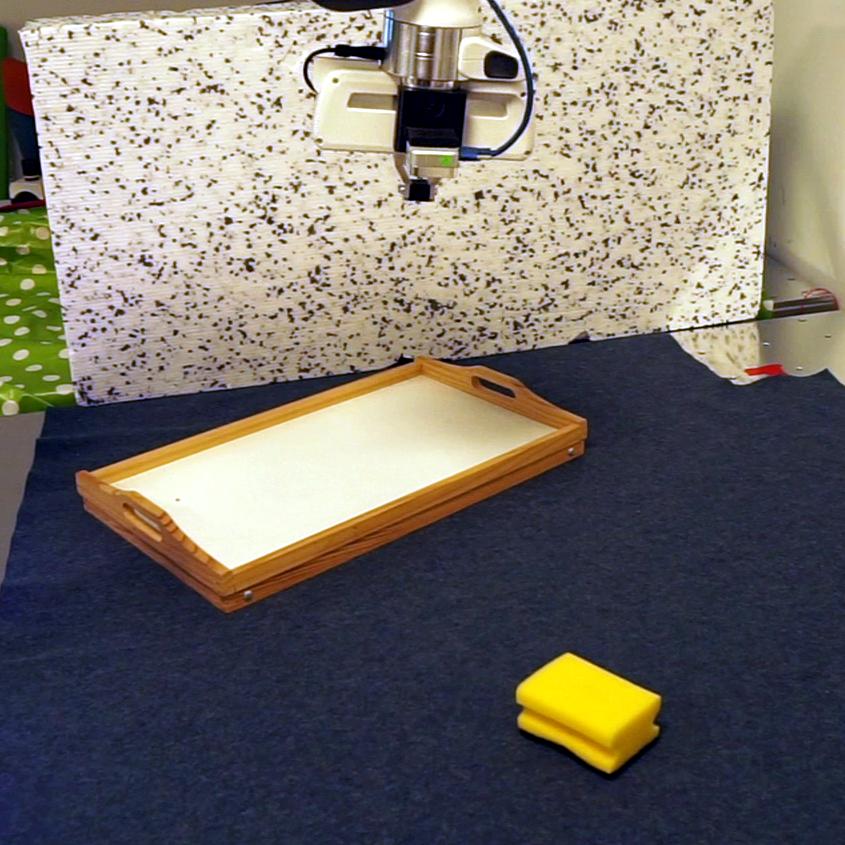}
        \caption{Failed Task Executions.}
        \label{subfig:real_world_failures}
    \end{subfigure}
    \caption{Real-World Robot Experiment. Failure cases include  imperfect grasping and premature closing off the gripper.}
    \label{fig:layout}
\end{figure}

We also evaluate the effectiveness of \ourName{} on a real-world robot system.
As done in PointFlowMatch~\cite{chisari2024learning}, we apply additional augmentations through transforming the input point cloud and proprioceptive states as well as the robot actions using a sampled $SE(3)$ transform. 
In \cref{sec:vlm_estimators}, we observed that the median error of the matching procedure scales objects 18\% too large. We incorporate this finding through downscaling all meshes by 0.8 for training data generation.
The remainder of the policy training for our real-world setup remains unchanged from our simulation experiments. For inference, we change the flow denoising steps to 30.
As our experimental platform, we use a Franka Panda robot with 7 degrees of freedom, shown in \cref{subfig:real_world_setup}. We attach a close-view Intel Realsense D405\footnote{We apply a manual filtering of the depth image when either close ($<5cm$) or far ($>50cm$).} to the wrist and use an Intel Realsense D435 as our external camera. We also align the simulation with our calibrated camera extrinsics.

We evaluate the \texttt{Sponge on Tray}-task on consecutive roll-outs for both 14 easy and 20 hard layouts. For the easier layouts, the average success rate achieved is $64\%$ and $30\%$ for the hard layouts. A successful execution is shown in \cref{subfig:real_world_success}.
The most common failure case is missing the sponge by just a few centimeters because the real-world depth camera becomes inaccurate for close-up manipulation. We present failures in \cref{subfig:real_world_failures}. For the \texttt{Coke on Tray} and \texttt{Paperroll upright}-task we achieve an average success rate of $21\%$ and $24\%$ respectively across 29 roll-outs each.
See the accompanying video for roll-outs of all tasks.\looseness=-1

\section{Conclusion}
\label{sec:conclusion}
In this work, we investigated the use of 3D generative foundational models to transfer a single human demonstration to simulation. We showed that with \ourName{}, more robust policies are learned with an increase in success rate of $\absoluteDeltaIncrease$ over baselines. We additionally analyzed the effort needed to generate task-relevant and usable simulation CAD models. We also showed the superior performance of our matching over VLM approaches and zero-shot transfer to a real-robot system.
While we showed the applicability of \ourName{} to generate expert demonstration data, it can also be used without any modifications to serve as a reinforcement learning environment. 

In the future, we see potential to extend \ourName{} to enable more tasks with constrained movements like interacting with articulated objects, e.g., sampled from an object prior like in CARTO~\cite{heppert2023carto}.
Another promising extensions could combine \ourName{} with the capabilities of VLMs as described in Gen2Sim~\cite{katara2024gen2sim} to setup tasks different from the human demonstration but using the same generated object meshes.

\FloatBarrier
\appendix
\subsection{Task Overview}
\label{app:sec:task_overview}
\newcommand{\notUsed}{No}

\begin{table}[h]
    \centering
    \begin{tabular}{rcccccc}
        \toprule
         &  
          \multicolumn{2}{c}{
                \makecell{\texttt{Sponge}\\\texttt{on Tray}}
            }
          & \multicolumn{2}{c}{
                \makecell{\texttt{Coke on}\\\texttt{Tray}}
            }
          & \multicolumn{2}{c}{
                \makecell{\texttt{Paperroll}\\\texttt{upright}}
            } \\
        &
          pos & rot & pos & rot & pos & rot \\
        \midrule
        \makecell[r]{Generation} 
          & 15cm & \notUsed 
          & 15cm & \notUsed 
          & 5cm & 10deg \\
        Evaluation 
          & 15cm & \notUsed 
          & 15cm & \notUsed 
          & \notUsed & 10deg \\
        \bottomrule
    \end{tabular}
    \caption{Overview of the Used Success Criteria.}
    \label{tab:task_success_criteria}
\end{table}
We evaluate \ourName{} on three different tasks, \texttt{Sponge on Tray}, \texttt{Coke on Tray} and \texttt{Paperroll upright}. Depending on the task we adapt the final success criterion in our demonstration generation and evaluation. \texttt{Sponge on Tray} and \texttt{Coke on Tray} both consist of grasping an object and placing it on a tray. For both tasks the poses of the objects are randomized and it is only important that the object is placed on the tray, thus we do not consider orientation. \texttt{Paperroll upright} differs as the task does not depend on a secondary object, but rather the paper roll has to be placed upright as the name suggests. Thus, during test time we only consider the orientation of the object but not the position. We given an overview of the criteria in 
\cref{tab:task_success_criteria}.

\subsection{Policy Hyper Parameters Setup}
\label{app:sec:policy_hps}
We adopt the training settings and hyperparameters from PointFlowMatch~\cite{chisari2024learning}. We randomly downsample the input point cloud to a size of 4096 points. The policy is trained using AdamW optimizer
with a learning rate of $3e^{-5}$ and weight decay of $1e^{-6}$. We schedule the learning rate with cosine annealing and linear warmup of 5000 steps. The batch size is 128, and we apply EMA on the weights of the model.
Throughout our experiments, we use $20$ denoising steps for flow matching and predict actions with a horizon of $\horizonLength = 32$.

\subsection{Real Objects Overview}
\begin{figure}[h]
    \centering
    \includegraphics[width=0.7\columnwidth]{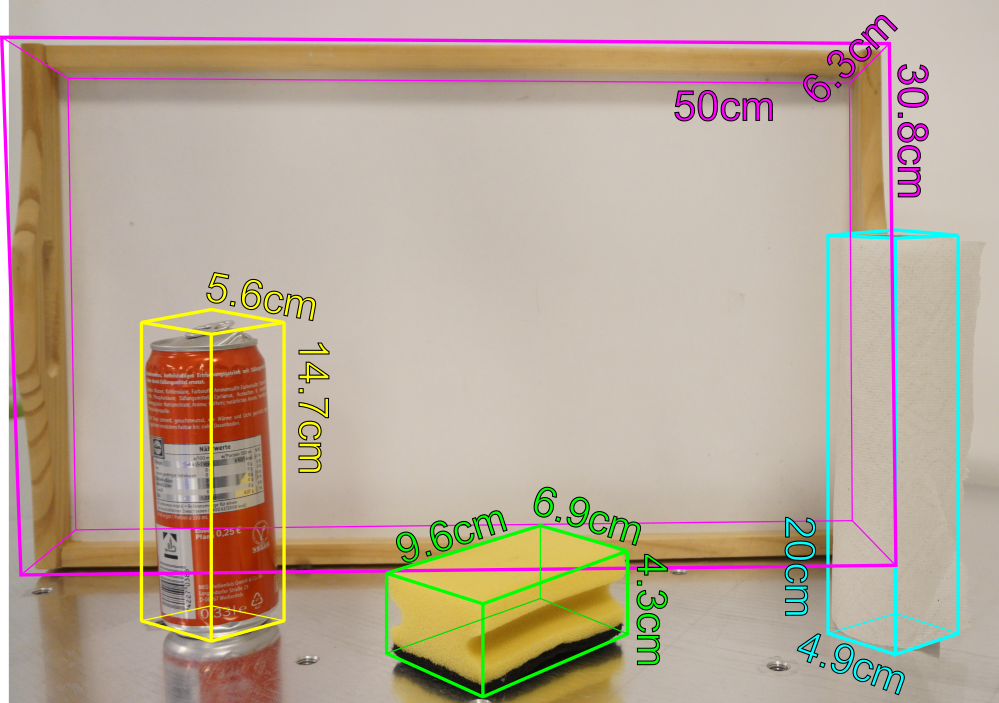}
    \caption{Real-World Reference Objects for Size Estimation.}
    \label{fig:real_object_sizes}
\end{figure}

For evaluating the dimensions in \cref{sec:vlm_estimators} and our real world robotic experiments in \cref{sec:exp:real}, we use the objects displayed in \cref{fig:real_object_sizes}.

\ifbool{fourPageVersion}{
    \subsection{Ablation Study}
    \label{app:sec:ablation}
    \begin{figure}
    \centering
    \includegraphics[width=0.8\linewidth]{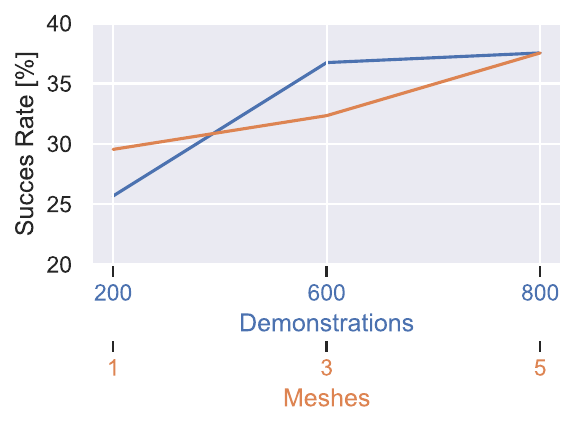}
    \caption{Results of Ablation Study. We show the average success rate $\lbrack \% \rbrack$ \mbox{($\uparrow$)} across all tasks. We either vary the number of demonstrations while using five meshes or we vary the number of meshes using 800 demonstrations.}
    \label{fig:ablation:demos_and_meshes}
\end{figure}

    In our ablation study, we investigate the effect of the amount of used meshes and the demonstration.
    Specifically, we change the number of demonstrations from 800 to 600 and 200, while still using five meshes, as well as changing the number of meshes to three and one with 800 demonstrations. We visualize the results in \cref{fig:ablation:demos_and_meshes}. As one would expect, the more meshes and demonstrations, the better. Nonetheless, the average performance increase diminishes rather soon when going from 600 to 800 ($1.1\%$) demonstrations and three to five meshes ($5.2\%$), respectively.
    
    \subsection{Extended Comparison of Mesh Generation}
    \label{app:sec:extend_mesh_generation}

    In \cref{tab:filter_effort} we show extended results from our mesh filtering experiment in \cref{sec:mesh_filtering}. We show results split by categories and individual pipeline steps.
}{
}
\FloatBarrier

\begin{footnotesize}
    \bibliographystyle{ieeetr}
    \bibliography{sources.bib}
\end{footnotesize}

\end{document}